%% file: main.tex
\documentclass[10pt,twocolumn,letterpaper]{article}

\usepackage[pagenumbers]{cvpr}
\usepackage{booktabs} 
\usepackage{float}    
\usepackage{bbding}
\usepackage{caption}  
\usepackage{fancyhdr}
\input{preamble}
\definecolor{cvprblue}{rgb}{0.21,0.49,0.74}
\usepackage[pagebackref,breaklinks,colorlinks,allcolors=cvprblue]{hyperref}
\usepackage{multirow}

\newlength{\yuvionheadextra}
\fancypagestyle{yuvionheader}{%
  \fancyhf{}%
  \fancyhead[L]{\raisebox{0pt}{\includegraphics[height=17pt]{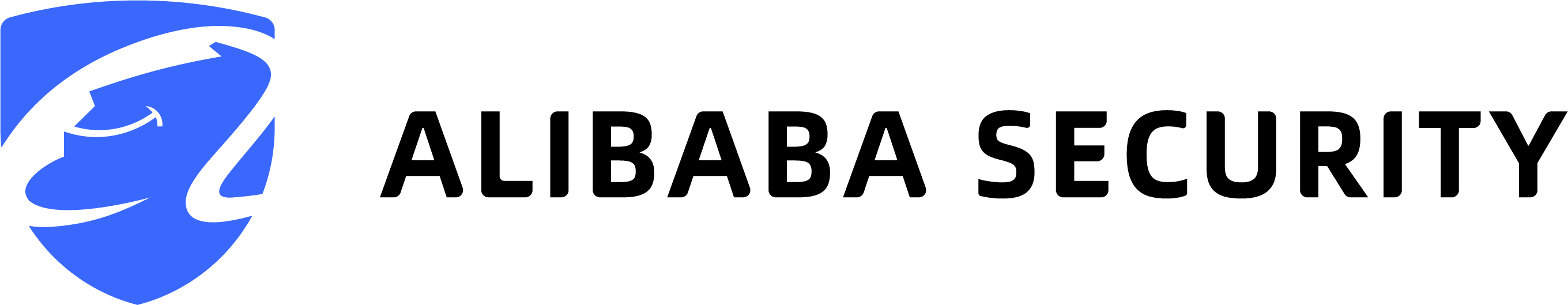}}}%
  \fancyhead[C]{}%
  \fancyhead[R]{\raisebox{0pt}{\includegraphics[height=17pt]{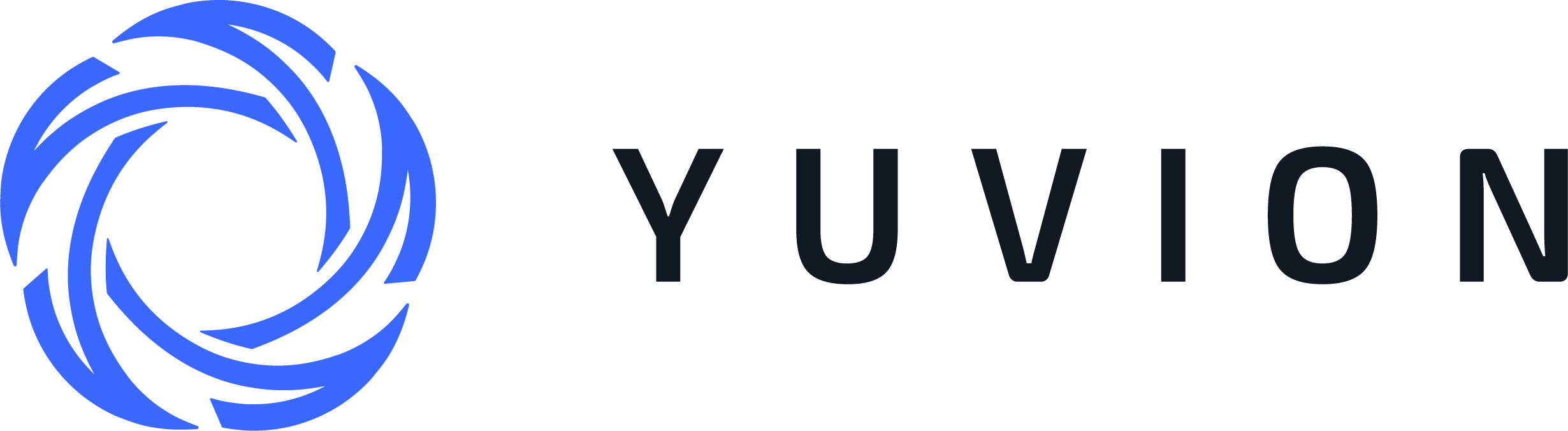}}}%
  \fancyfoot[C]{\thepage}%
}
\def\paperID{*****} 
\def\confName{CVPR}
\def\confYear{2026}

\title{Diffusion Probe: Generated Image Result Prediction Using CNN Probes}

\author{
    Benlei Cui$^{1,*}$ \quad Bukun Huang$^{2,*}$ \quad Zhizeng Ye$^{2}$ \quad Xuemei Dong$^{2,\text{\protect\Envelope}}$ \quad Tuo Chen$^{3}$ \\
    Hui Xue$^{1}$ \quad Dingkang Yang$^{4,5}$ \quad Longtao Huang$^{1}$ \quad Jingqun Tang$^{5}$ \quad Haiwen Hong$^{1,\ddagger,\text{\protect\Envelope}}$ \\
    \noalign{\addvspace{8pt}}
    $^{1}$Yuvion Team, Alibaba Group \\ $^{2}$Laboratory for Statistical Monitoring and Intelligent Governance of Common Prosperity,\\ School of Statistics and Data Science, Zhejiang Gongshang University \\ $^{3}$Southeast University \\
    $^{4}$College of Intelligent Robotics and Advanced Manufacturing, Fudan University \\ $^{5}$ByteDance Inc. \\
    \noalign{\addvspace{5pt}}
    {\tt\small \{cuibenlei.cbl, hui.xueh, kaiyang.hlt, honghaiwen.hhw\}@alibaba-inc.com} \\
    {\tt\small \{23020040119, 2002090138\}@pop.zjgsu.edu.cn, dongxuemei@zjgsu.edu.cn}\\
    {\tt\small tchen@seu.edu.cn \quad dicken@fysics.ai \quad jingquntang@163.com}
}
\begin{document}
\maketitle
\linespread{0.99}\selectfont
\pagestyle{yuvionheader}
\thispagestyle{yuvionheader}
\begingroup
\renewcommand{\thefootnote}{}
\footnotetext{%
\begin{minipage}[t]{\linewidth}
\setlength{\parindent}{0pt}
\begin{tabular*}{\linewidth}{@{\extracolsep{\fill}}lll@{}}
$^*$ Equal contribution. & $^\ddagger$ Project leader. & $^\text{\protect\Envelope}$ Corresponding author.
\end{tabular*}
\par\vspace{1pt}
\textbf{Accepted to CVPR 2026.} Benlei Cui conceived the core idea, completed the manuscript writing, and conducted part of the experiments. Bukun Huang completed the main experiments and contributed to part of the manuscript writing.
\end{minipage}}
\endgroup
\input{sec/0_abstract}    
\input{sec/1_intro}
\input{sec/2_relatedwork}
\input{sec/3_method}

\input{sec/4_exp}

\clearpage
\linespread{1}\selectfont
{
    \scriptsize
    \bibliographystyle{ieeenat_fullname}
    \bibliography{main}
}

\input{sec/X_suppl}
\end{document}

%% file: sec/0_abstract.tex
\begin{abstract}
Text-to-image (T2I) diffusion models currently lack an efficient mechanism for early quality assessment, forcing costly random trial-and-error in scenarios requiring multiple generations (e.g., iterating on prompts, agent-based image generation, flow-grpo). To address this, we first reveal a strong correlation between the attention distribution in the early diffusion process and the final image quality. Building upon this insight, we introduce \textbf{Diffusion Probe}, a pioneering framework that leverages the model’s internal cross-attention maps as a predictive signal. We propose a lightweight predictor, trained to establish a direct mapping from statistical properties of these nascent cross-attention distributions—extracted from the initial denoising steps—to the final image’s comprehensive quality. This allows our probe to accurately forecast various aspects of image quality, regardless of the specific ground-truth quality metric, long before full synthesis is complete.
We empirically validate the reliability and generalizability of Diffusion Probe through its consistently strong predictive accuracy across a wide spectrum of conditions. On diverse T2I models, throughout broad early-denoising windows, across various resolutions, and with different quality metrics, it achieves high correlation (PCC $>$ 0.7) and classification performance (AUC-ROC $>$ 0.9). This intrinsic reliability is further demonstrated in practice by successfully optimizing T2I workflows that benefit from early, quality-guided decisions, such as \textbf{Prompt Optimization}, \textbf{Seed Selection}, and \textbf{Accelerated RL Training}.
In these applications, the probe's early signal enables more targeted sampling strategies, preempting costly computations on low-potential paths. This yields a dual benefit: a significant reduction in computational overhead and a simultaneous improvement in final outcome quality, establishing Diffusion Probe as a model-agnostic and broadly applicable tool poised to revolutionize T2I efficiency. Our code is available at  \href{https://github.com/Alibaba-VELLDEPTH/DiffusionProbe}{https://github.com/Alibaba-VELLDEPTH/DiffusionProbe}.

\end{abstract}

%% file: sec/1_intro.tex
\section{Introduction}
\label{sec:intro}

The rapid advancements in diffusion-based text-to-image (T2I) generation models~\cite{flux2024,wu2025qwenimagetechnicalreport} have revolutionized visual content creation, enabling the synthesis of high-fidelity images directly from natural language descriptions. Despite their remarkable success, T2I models still face challenges in consistently generating images that perfectly align with complex or detailed prompts. Common issues include object distortion, omission, or semantic misalignment, which compromise aesthetic quality and practical utility. To address these shortcomings, an \textbf{iterative resampling process} is present in both practical applications and academic research. This is evident in users iteratively refining prompts to achieve desired outcomes. In academic methods, IC-Edit~\cite{zhang2025ICEdit} employs repeated generation to adjust output quality, while reinforcement learning frameworks like Flow-GRPO~\cite{liu2025flow} generate multiple results from the same prompt to construct relative ranking losses. Similarly, agent-based systems~\cite{chen2025t2icopilottrainingfreemultiagenttexttoimage} also rely on iterative sampling to progressively adjust their results.

\begin{figure}
    \centering
    \includegraphics[width=\linewidth]{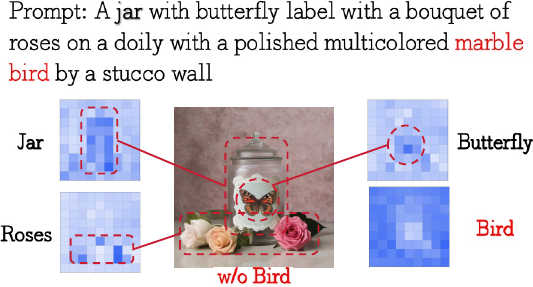} 
    \caption{Illustration of early cross-attention dispersion. Here, we present the prompt, the corresponding four cross-attention activation maps in the early denoising stage, and the final generated image. Compared to other tokens, the cross-attention activation maps of the ``bird'' token shows significant sparsity in spatial distribution.}
    \label{fig:bird_example_a}
\end{figure}

A core limitation inherent to these existing enhancement methods is that they often require completing the entire denoising process. This necessitates significant computational resources and time, especially when dealing with a large search space or numerous iterations to find optimal configurations. Consequently, these methods, while demonstrably effective, frequently incur prohibitive costs, thereby hindering their widespread adoption and scalability. This highlights a critical need for more efficient diagnostic and predictive mechanisms capable of probing the final generation quality at the early stages of the diffusion process.

Extending the paradigm of probing techniques from Large Language Models (LLMs)~\cite{mckenzie2025detectinghighstakesinteractionsactivation} to the domain of text-to-image (T2I) generation, we investigate the relationship between early-stage cross-attention maps and final image quality. Prior studies have affirmed the predictability of final outcomes from early generative stages. For instance, ICEDIT~\cite{zhang2025ICEdit} forecasts quality by decoding an early-stage latent for evaluation by a Vision-Language Model (VLM). This approach, however, incurs substantial computational overhead due to the reliance on an external 72b VLM. In parallel, the diagnostic utility of cross-attention maps has been acknowledged. PromptCharm~\cite{wang2024promptcharm}, for example, visualizes these maps to provide qualitative user feedback, but its reliance on human interpretation precludes its use in automated, quantitative pipelines.

While these methods validate a link between early generative features and final quality, they are limited by either high computational costs or a lack of automation. Accordingly, our research is predicated on the hypothesis that the raw numerical patterns within early-stage cross-attention maps can serve as a direct, lightweight proxy for eventual image quality, thereby circumventing both expensive decoding and manual analysis. Our empirical analysis validates this hypothesis, revealing a strong correlation: fragmented and diffuse cross-attention patterns are highly predictive of generation failures, such as object omissions or semantic inconsistencies. As depicted in \autoref{fig:bird_example_a}, a dispersed attention map for the “bird” token during early denoising stages directly foreshadows its poor rendition in the final image. These findings establish early-stage cross-attention as a potent and computationally efficient \emph{probe} for assessing image fidelity.

Therefore, we introduce Diffusion Probe, a foundational framework that, for the first time, systematically quantifies and operationalizes the link between early-stage cross-attention and final image quality. Diffusion Probe transcends this limitation by using a lightweight yet powerful CNN probe trained to map nascent attention patterns to any quantifiable image attribute, such as aesthetic scores, semantic accuracy, or object fidelity. Crucially, it decouples quality prediction from the image synthesis pipeline, providing an accurate forecast without incurring the computational cost of either full image generation or post-hoc evaluation.

The implications of this direct, early-stage quality prediction are far-reaching. Diffusion Probe provides a universal guidance mechanism applicable to virtually any optimization task within T2I generation that relies on evaluating multiple candidates—a process traditionally hampered by the high cost of full image generation. This paradigm shift unlocks fundamental improvements in both efficiency and performance across a spectrum of applications. We demonstrate its transformative potential in three exemplary domains: (1) Automated Prompt Optimization, by rapidly iterating through prompt variations to find optimal phrasing; (2) Efficient Seed Selection, by preemptively discarding unpromising generation trajectories; and (3) Accelerated Policy Learning, by providing a dense and computationally cheap reward signal for reinforcement learning. These applications merely scratch the surface of Diffusion Probe’s potential, as it offers a versatile building block for future research in controllable and efficient T2I synthesis.

Our contributions are as follows:  
\begin{enumerate}[leftmargin=*]
\item By introducing the concept of a probe to diffusion models for the first time, we reveal a fundamental insight: that the complex final quality of a text-to-image (T2I) generation is predictably encoded within its early-stage cross-attention patterns. This establishes cross-attention as a powerful, emergent diagnostic signal for anticipating generative trajectories, enabling proactive assessment without costly full-generation rollouts.

\item Based on this insight, we introduce \textbf{Diffusion Probe}, a novel and lightweight framework that leverages early attention patterns for quality prediction. We empirically validate its robustness as a model-agnostic tool by showing it achieves high predictive accuracy—with SRCC exceeding \textbf{0.8} and AUC surpassing \textbf{0.9}—across diverse T2I models like the UNet-based SDXL and the DiT-based FLUX.1 and Qwen-Image.

\item We demonstrate Diffusion Probe's significant practical impact in accelerating and optimizing T2I workflows that necessitate multiple sampling steps. We exemplify its utility in diverse applications, including efficient \textbf{prompt optimization}, cost-effective \textbf{seed selection}, and significantly accelerated training convergence for RL-based generative policies like \textbf{Flow-GRPO}.
\end{enumerate}

%% file: sec/2_relatedwork.tex
\section{Related Work}
\label{sec:related work}

\textbf{Text-to-image Generation.} 
Text-to-image (T2I) generation has been revolutionized by diffusion models (DMs) \cite{ho2020denoising, song2021scorebased}, now the dominant paradigm producing images of unprecedented quality and diversity. Key milestones include large-scale models like Imagen \cite{saharia2022photorealistic}, DALL-E 3 ~\cite{betker2023Dalle3}, and the widely adopted open-source Stable Diffusion family \cite{rombach2022high}. Recent advancements with Diffusion Transformer (DiT) architectures, such as FLUX ~\cite{flux2024} and Qwen-Image ~\cite{wu2025qwenimagetechnicalreport}, have further pushed performance boundaries. Beyond generic image synthesis, diffusion models have been widely applied to object removal, product-poster generation, virtual try-on, and efficient image and video sampling~\cite{sun2025attentiveeraser,liu2025erasediffusion,cui2026tcpade,huang2026curvatureaware,cui2026simpleposter,li2025shoefit}. Despite these successes, T2I models still face challenges in reliability and efficiency, especially with complex prompts. Our work addresses these limitations by leveraging early-stage cross-attention patterns to improve generation outcomes.

\noindent\textbf{Enhancing T2I Generation Quality.} 
The computational cost of achieving high-quality T2I generation for complex prompts has spurred various optimization strategies. \textbf{Prompt optimization} techniques refine textual inputs, evolving from manual tuning or iterative searching~\cite{hertz2022prompt} to automated rewriting using large language models (LLMs)~\cite{promptist,PromptCrafter,wang2024promptcharm}. \textbf{Seed selection and reranking} methods generate multiple candidates to select the best, sometimes using early-stage feature decoding for more efficient selection; ICEDIT~\cite{zhang2025ICEdit}, for instance, uses this for image editing. Concurrently, \textbf{reinforcement learning (RL) paradigms} like DDPO~\cite{black2023training} and Flow-GRPO~\cite{liu2025flow} have successfully fine-tuned diffusion models for improved alignment and aesthetics. However, a critical limitation unites these approaches: their reliance on evaluating complete generation outputs. This incurs high computational costs from repeated denoising, whether for prompt refinement, seed selection, or RL reward collection. Our work circumvents this by enabling a proactive, early-stage quality assessment without requiring full generation, thus providing a more efficient foundation for these optimization tasks.

\noindent\textbf{Attention Mechanisms in T2I Models.} 
Inspired by the burgeoning field of \textbf{LLM probing}~\cite{alain2018understandingintermediatelayersusing,mckenzie2025detectinghighstakesinteractionsactivation, marks2024geometrytruthemergentlinear}, which analyzes internal model states to understand and predict model behavior, our work extends this diagnostic paradigm to T2I models. In LLMs, probing tasks involve training simple "probe" models on representations extracted from pre-trained LLMs to predict linguistic properties (e.g., part-of-speech tags, syntactic trees, semantic roles) \cite{hewitt2019structural}. Classic works such as \cite{conneau2018cramsinglevectorprobing, tenney2019bertrediscoversclassicalnlp} established that LLMs encode rich linguistic information within their internal layers, often discernible through their attention mechanisms ~\cite{clark2019doesbertlookat}. This allows researchers to diagnose model capabilities, identify biases, and forecast task performance (e.g., identifying failure modes or predicting fine-tuning success) without full model execution.

In Diffusion Models, attention mechanisms, particularly cross-attention, are similarly fundamental for aligning text prompts with visual features. Consequently, a substantial body of research has focused on analyzing and manipulating these attention patterns for improved control and fidelity. Works like DAAM \cite{tang2023daam} study the role of cross-attention in semantic analysis and interpretability by producing pixel-level attribution maps. Other methods aim to directly manipulate attention during inference for better image generation. Unlike interventional methods that actively guide attention during generation to improve quality or control (e.g., SAG~\cite{hong2022improving}, Attend-and-Excite~\cite{chefer2023attendandexcite}), our approach is non-invasive and predictive. We establish a direct link between \emph{early-stage attention patterns} and final image quality. This serves as an early-stage diagnostic tool for downstream tasks without modifying the generation process.

%% file: sec/3_method.tex
\section{Methodology}

\subsection{Preliminaries}

\textbf{Diffusion Models.}
Diffusion Models (DMs)~\cite{ho2020denoising} are generative models that learn to reverse a fixed Markovian process that progressively adds noise to a data sample $\mathbf{x}_0$ over $T$ timesteps. The model's core is a denoising network, $\epsilon_\theta(\mathbf{x}_t, t, c)$, trained to predict the added noise in a corrupted sample $\mathbf{x}_t$ given the timestep $t$ and context $c$. To generate a sample, this network is iteratively applied starting from pure noise $\mathbf{x}_T$, progressively denoising it to align with the learned data distribution.

\begin{figure*}
    \centering
    \includegraphics[width=\linewidth]{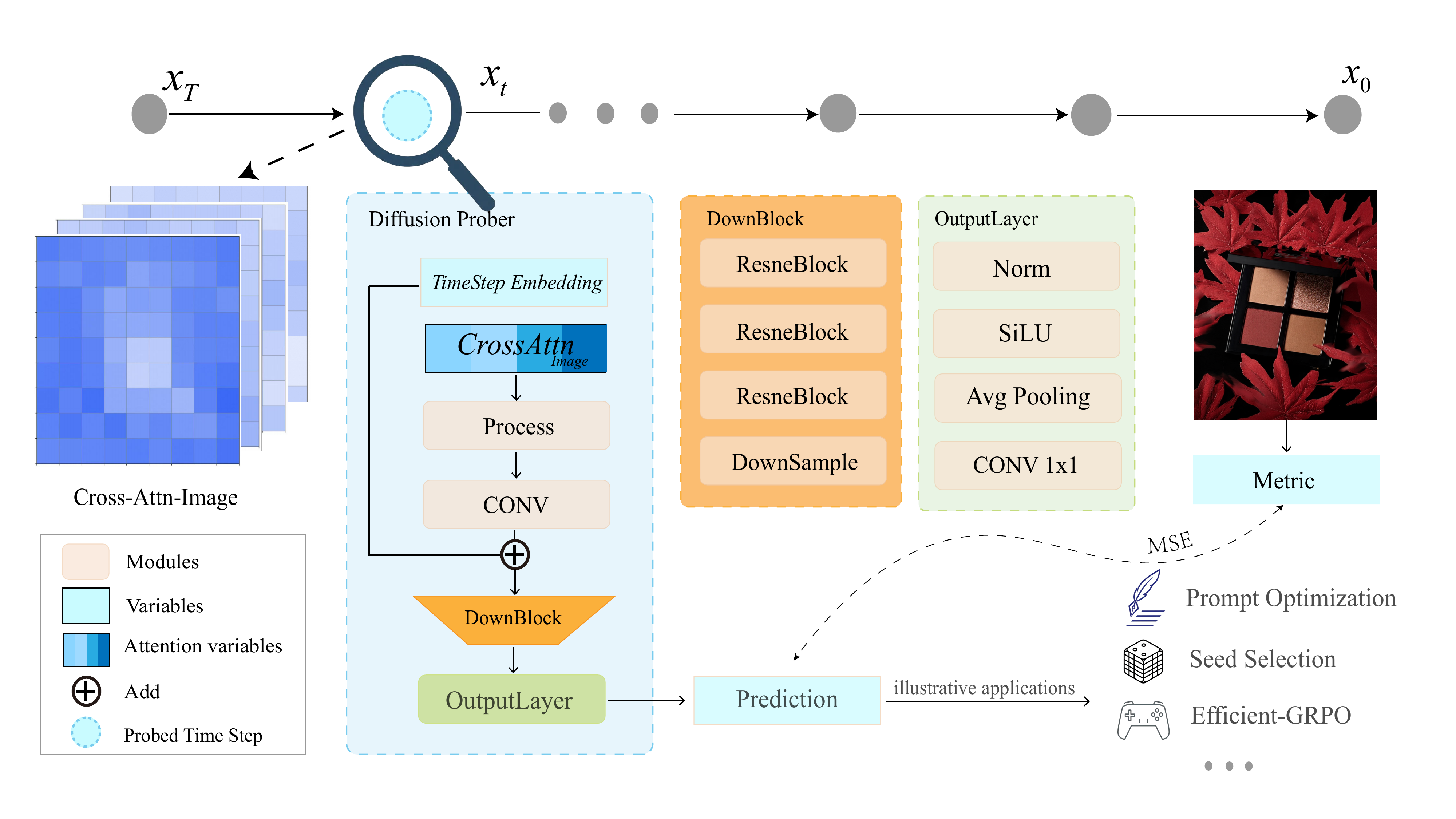}
    \caption{
        \textbf{Overview of the Diffusion Probe framework.} Our framework takes as input the early-stage cross-attention feature maps (derived from the \textit{CrossAttn} module at a probed timestep $t$) and the \textit{TimeStep Embedding}. A lightweight network processes these inputs, ultimately outputting a quality score prediction for the final generated image ($x_0$). This predicted score is learned to align with a specified ground-truth \textit{Metric} (e.g., aesthetic, semantic coherence) evaluated on the fully synthesized image. The Diffusion Probe then serves as a versatile tool to enable various downstream applications, such as Prompt Optimization, Seed Selection, and Efficient-GRPO training.
    }
    \label{fig:train_prober}
\end{figure*}

\textbf{Multimodal Diffusion Transformers (MM-DiT).}
Modern text-to-image models increasingly favor Diffusion Transformer (DiT) backbones over U-Nets~\cite{DBLP:journals/corr/RonnebergerFB15} for their superior scalability. Our work builds on the Multimodal Diffusion Transformer (MM-DiT)~\cite{esser2024scalingrectifiedflowtransformers}, a notable DiT variant. A key architectural feature of MM-DiT is its modality fusion. Instead of cross-attention, it concatenates the image latent $\mathbf{z}_{\text{img}}$ and text latent $\mathbf{z}_{\text{text}}$ into a single sequence: $\mathbf{h} = [\mathbf{z}_{\text{text}}; \mathbf{z}_{\text{img}}]$.
\emph{Crucially}, a single self-attention mechanism operates on this unified representation $\mathbf{h}$, enabling direct and deep interaction between textual and visual tokens within the same attention operation. This architectural transparency, where cross-modal interactions are explicitly captured in self-attention patterns, makes MM-DiT an ideal testbed for our work, which probes these patterns to forecast generation quality.

\subsection{Attention-Quality Mapping in MMDiTs}

Although diffusion models exhibit strong performance in image synthesis, the quality of any given sample is difficult to predict a prior. Consequently, users often adopt iterative sampling—repeatedly adjusting seeds and prompt formulations—to obtain a satisfactory image. This reliance on exploratory resampling remains prevalent even in state-of-the-art models such as FLUX.

\textbf{Core insight.} In diffusion-based text-to-image (T2I) models, initial denoising steps primarily establish \emph{global structural coherence and coarse spatial layout}, with later steps refining \emph{local attributes}. Our core insight identifies cross-attention mechanisms as a transparent and early diagnostic probe into this generative process. We observe that for text tokens, cross-attention maps rapidly form compact, stable spatial focus, indicating early object grounding. This makes cross-attention an ideal signal to predict final image quality without full synthesis.

\begin{figure}[t]
    \centering
    \includegraphics[width=\linewidth]{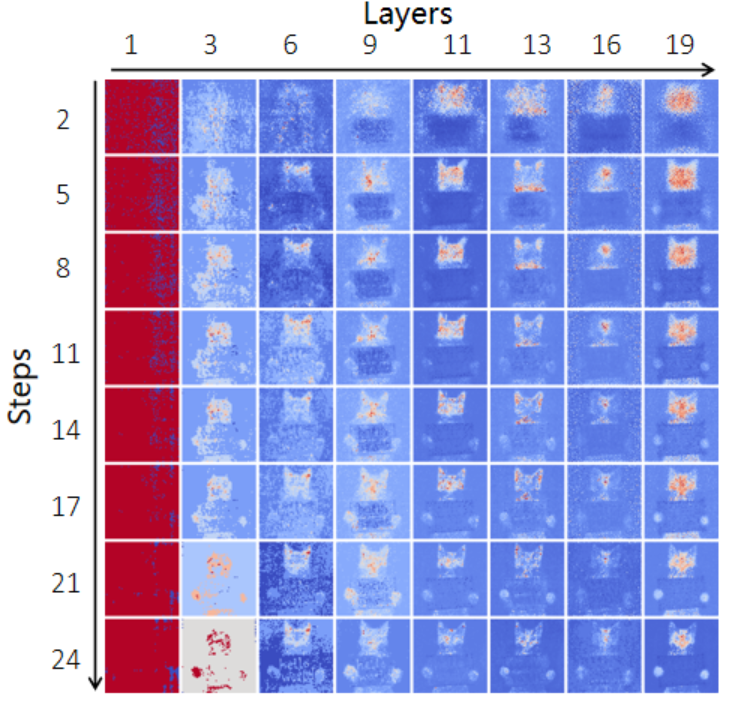}
    \caption{
        Visualization of the Cross-Attention Map for the token "cat" within a FluxTransformerBlock. The image illustrates the spatial attention distribution for the text token \textit{"cat"} generated from the prompt ``A cat holding a sign that says hello world''. Regions with intense red patterns indicate high attention scores, demonstrating where the model's focus is directed in response to the specified token. 
    }
    \label{fig:fluxvisual}
\end{figure}

\begin{figure}
    \centering
    \includegraphics[width=1\linewidth]{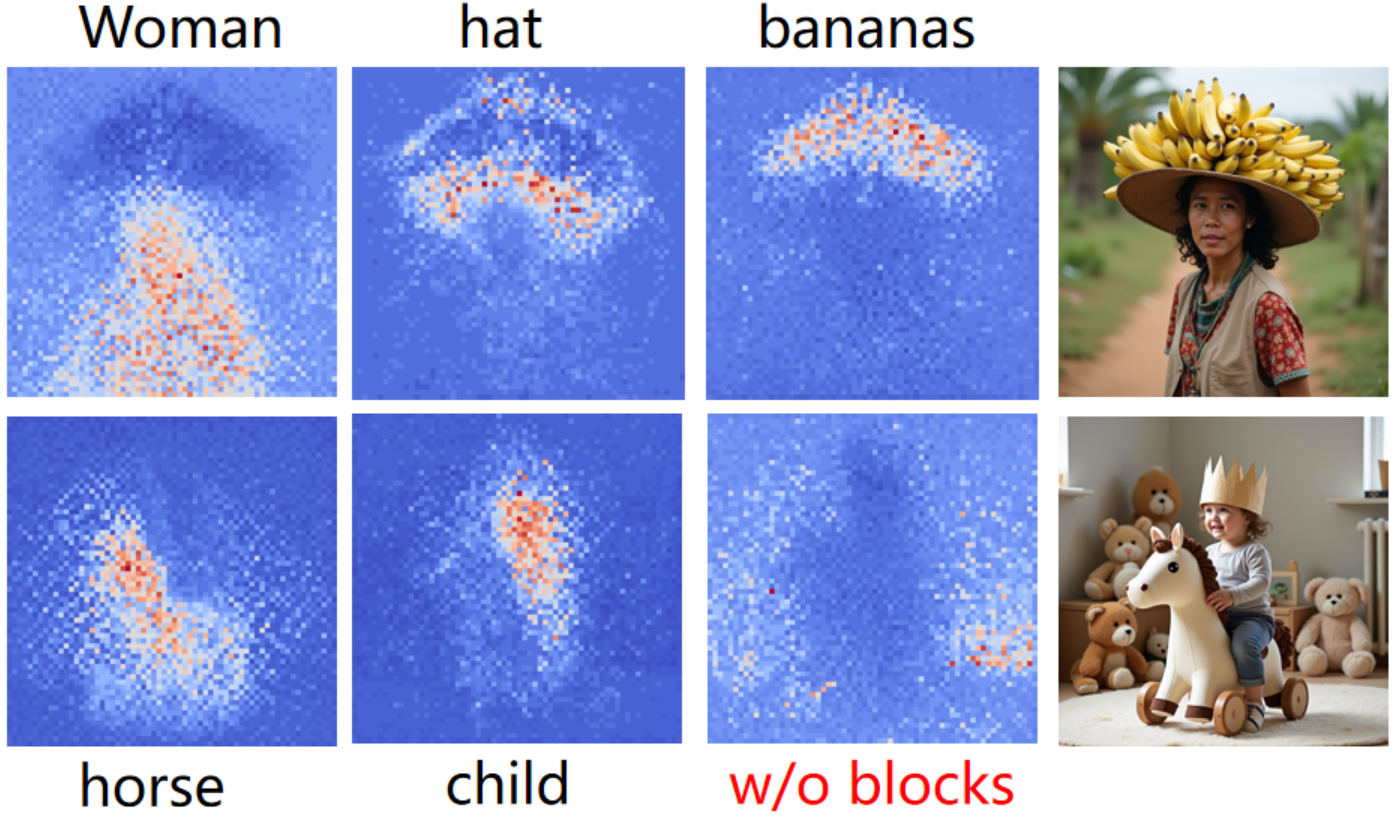}
    \caption{Early-stage cross-attention maps reveal object rendering fidelity. \textbf{(Top)} For the prompt ``Woman carrying a bunch of bananas on top of her hat'', the model successfully renders all objects, resulting in sharp, focused attention maps for each token. \textbf{(Bottom)} In contrast, for ``A child ... surrounded by ... building blocks in a playroom'', the model fails to generate the \textbf{building blocks}, and the corresponding attention map becomes highly diffuse. This demonstrates that attention statistics serve as a reliable early indicator of object-level generation success or failure (maps extracted at step $t=5$, layer 19, as detailed in Figure~\ref{fig:fluxvisual}).}
    \label{fig:attentionmap}
\end{figure}

\textbf{Our Evidence.} We audit FLUX by visualizing cross-attention across timesteps $t$ and DiT blocks $b$. We observe two critical phenomena, each supported by distinct visual evidence.

\emph{Firstly}, as exemplified in Figure~\ref{fig:fluxvisual}, even within the high-noise denoising regime, semantically salient object tokens reliably induce sharply defined, localized regions of high attention. These nascent attentional foci delineate a rudimentary spatial outline, confirming rapid initial object grounding. This early focusing behavior is stable across initial timesteps and progressively sharpens in deeper DiT blocks.

\emph{Second}, and crucially, as depicted in Figure~\ref{fig:attentionmap}, whenever the final generated image exhibits low quality or semantic failures (e.g., missing objects, distortions, and attribute mismatches), the corresponding early attention patterns are visibly \emph{diffuse and fragmented}. Instead of forming a coherent focus, the attention maps spread over multiple irrelevant regions or oscillate unstably across denoising steps. This fragmentation directly correlates with poor outcomes, while concentrated and stable early attention precedes high-fidelity generations. We provide an expanded gallery of such cases, covering a diverse range of failure modes, in Appendix~\ref{sec:appendix_failure}.

These two observations—early outline formation and early-time dispersion in failure cases—establish cross-attention as a potent \emph{probe} of eventual fidelity. This directly motivates our approach: a predictor that learns to extract an image-level quality signal from these nascent attention patterns, anticipating final outcomes without completing the full denoising trajectory.

\subsection{Predicting Image Quality from Cross-Attention}

We introduce \textbf{Diffusion Probe}, a lightweight predictive framework designed to forecast final image quality by inspecting the cross-attention mechanisms in the initial stages of the generative process. As illustrated in Figure~\ref{fig:train_prober}, our approach is implemented via a supervised model, which avoids costly full-generation rollouts by learning to interpret early signals of the model's generative trajectory.

Formally, for a given text prompt and a denoising step $t \in \{1, \dots, T_0\}$, we extract the cross-attention maps of each words in the prompts, denoted as the set $\mathcal{A}$. Our extraction strategy is designed to be architecture-agnostic, targeting the middle stages of the model's encoding path. This applies to both traditional UNet-based models and modern transformer-based architectures like MMDiT (e.g., FLUX). We specifically chose this intermediate stage because, as visually evidenced in Figure~\ref{fig:fluxvisual}, the cross-attention maps at this point exhibit a distinct semantic structure and spatial layout. This provides a rich, interpretable signal of text-image alignment before the features become overly compressed in deeper layers.

The core of our framework is the Probe, $E_\theta$, which processes these specific attention maps $\mathcal{A}$ along with the corresponding TimeStep Embedding to project the high-dimensional attention data into a compact latent representation. As depicted in the diagram, the Diffusion Probe's architecture is composed of several DownBlocks with residual layers and an OutputLayer that uses normalization, pooling, and convolutions to produce the final scalar prediction.

This entire network, including the final prediction head $f_\theta$, is trained end-to-end to map the latent representation to a scalar quality score:
\[
\hat{q} = f_\theta(E_\theta(\mathcal{A}, t)).
\]
The probe is trained on an offline dataset using a straightforward regression objective. For each generated image, we obtain a scalar quality score $q$ from a pre-trained reward model (e.g., an aesthetics predictor), which serves as the ground truth label (Metric in Figure~\ref{fig:train_prober}). The training objective is to minimize the Mean Squared Error (MSE) between the probe's predicted score $\hat{q}$ and the ground truth score $q$. The loss function is defined as:
\[
\mathcal{L}=\lVert \hat{q}-q\rVert_2^2.
\]
This MSE loss drives the model to predict the absolute quality score accurately by penalizing the squared difference between the prediction $\hat{g}$ and the ground truth $q$. This simple yet effective objective trains the probe to serve as a reliable estimator of the final image's quality.

At inference, the predicted score $\hat{q}$ acts as a potent, early-stage signal for guiding the generation process. This score essentially functions as an efficient \textit{probe} into the generative trajectory, leveraging early-stage cross-attention alignments to forecast the final output quality. The primary utility of this early prediction lies in its ability to dramatically accelerate workflows that rely on trial-and-error sampling. By accurately identifying promising or flawed generative paths within the first few steps, Diffusion Probe offers a computationally inexpensive mechanism to distinguish between high-quality and low-quality results long before the full sampling process is complete. Crucially, our approach operates as a plug-and-play module without requiring any modification to the pretrained foundation model, providing a universal and efficient tool to accelerate generative workflows.

\subsection{Downstream Applications}
We demonstrate Diffusion Probe's utility in three applications, where its early quality prediction $\hat{q}$ is leveraged to improve generation quality and efficiency.

\paragraph{Predictive Prompt Optimization.}
The probe acts as an efficient gatekeeper. For a given prompt $p$, we compute its predicted quality $\hat{q}(A(p, s))$. If the score falls below a threshold $\tau$, the prompt is selectively forwarded to a Large Language Model (LLM) for refinement. This preempts poor generations and incurs LLM costs only when necessary. The process is formalized as:
\[
\tilde{p} = 
\begin{cases} 
    \mathrm{LLM}(p) & \text{if } \hat{q}(A(p, s)) < \tau \\
    p & \text{otherwise}
\end{cases}
\]

\paragraph{Efficient Seed Selection.}
To find the optimal seed from a large pool $\mathcal{S}$ for a fixed prompt, we generate partial trajectories for each seed (for only $T_0 \ll T$ steps) and use the probe to predict their final quality. The seed yielding the highest predicted score is then selected for the single, full generation run. This efficiently filters out low-potential seeds, replacing multiple costly full generations with one informed choice.
\[
s^\star = \operatorname*{argmax}_{s \in \mathcal{S}} \hat{q}(A(p, s)).
\]

\paragraph{Efficient Flow-GRPO Training.}
The probe's prediction $\hat{q}$ serves as a low-cost, early-stage reward signal for methods like Flow-GRPO. It enables rapid mining of preference pairs $(x^+, x^-)$ by partitioning early-stage trajectories into positive ($\mathcal{D}^+$) and negative ($\mathcal{D}^-$) sets based on a quality threshold. This bypasses the need for costly full rollouts to find suitable training data, significantly accelerating policy convergence.

\vspace{1em} 
Across these applications, Diffusion Probe functions as a model-agnostic module that forecasts final quality from early-stage signals. This allows for the pruning of low-potential generative paths, concentrating compute on promising candidates to enhance both efficiency and output quality without modifying the base model.

%% file: sec/4_exp.tex
\section{Experiments}

\begin{table*}[h!]
    \centering
    \small 
    \caption{
        Diffusion Probe's Predictive Accuracy for External Image Quality Metrics across Diffusion Steps (at 1024$\times$1024 Resolution).
        This table evaluates the predictive accuracy of our Diffusion Probe's internal scores in aligning with established external image quality metrics. We quantify prediction performance using four distinct rank-based and correlation metrics. The Step column indicates the diffusion step from which the probe's attention features were extracted for prediction. Higher values indicate superior predictive alignment.
    }
    \label{tab:comprehensive_results}
    \setlength{\tabcolsep}{4pt} 
    \begin{tabular}{l l c | c c c c} 
        \toprule
        \textbf{Base Model} & \textbf{Resolution} & \textbf{Step} & \textbf{SRCC $\uparrow$} & \textbf{AUC-ROC $\uparrow$} & \textbf{KTC $\uparrow$} & \textbf{PCC $\uparrow$} \\
        \midrule
        \multirow{4}{*}{\textbf{SDXL~\cite{podell2023sdxl}}} & \multirow{4}{*}{1024$\times$1024} & 1  & 0.49 & 0.53 & 0.35 & 0.48 \\
                                       &                                  & 5  & 0.73 & 0.86 & 0.57 & 0.72 \\
                                       &                                  & 10 & \textbf{0.76} & \textbf{0.89} & \textbf{0.61} & \textbf{0.75} \\
                                       &                                  & 15 & 0.75 & 0.89 & 0.60 & 0.74 \\
        \midrule
        \multirow{4}{*}{\textbf{FLUX~\cite{flux2024}}} & \multirow{4}{*}{1024$\times$1024} & 1  & 0.52 & 0.62 & 0.38 & 0.50 \\
                                       &                                  & 5  & 0.76 & 0.88 & 0.60 & 0.75 \\
                                       &                                  & 10 & \textbf{0.79} & \textbf{0.91} & \textbf{0.64} & \textbf{0.78} \\
                                       &                                  & 15 & 0.78 & 0.91 & 0.63 & 0.77 \\
        \midrule
        \multirow{4}{*}{\textbf{Qwen-Image~\cite{wu2025qwenimagetechnicalreport}}} & \multirow{4}{*}{1024$\times$1024} & 1  & 0.45 & 0.67 & 0.32 & 0.44 \\
                                       &                                  & 5  & 0.69 & 0.84 & 0.53 & 0.68 \\
                                       &                                  & 10 & \textbf{0.72} & \textbf{0.87} & \textbf{0.56} & \textbf{0.71} \\
                                       &                                  & 15 & 0.71 & 0.86 & 0.55 & 0.70 \\
        \bottomrule
    \end{tabular}
\end{table*}

\begin{table*}[h!]
    \centering
    \caption{Performance comparison of Diffusion Probe in Prompt Optimization and Seed Selection tasks on SDXL and FLUX models. All metrics are reported as higher is better ($\uparrow$).}
    \label{tab:combined_results}
    \begin{tabular}{lll ccc}
        \toprule
        \textbf{Model} & \textbf{Task} & \textbf{Method} & \textbf{CLIP Score $\uparrow$} & \textbf{ImageReward $\uparrow$} & \textbf{Aesthetic Score $\uparrow$} \\
        \midrule
        \multirow{5}{*}{SDXL} & \multirow{3}{*}{Prompt Opt.} & Baseline & 28.31 & 0.71 & 5.13 \\
        & & + Diffusion Probe (Ours) & 30.24 & 0.72 & 5.29 \\
        & & + LLM & \textbf{30.80} & \textbf{0.73} & \textbf{5.34} \\
        \cmidrule(lr){2-6}
        & \multirow{2}{*}{Seed Sel.} & Baseline (Random) & 28.31 & 0.71 & 5.13 \\
        & & + Diffusion Probe (Ours) & \textbf{30.14} & \textbf{0.82} & \textbf{5.43} \\
        \midrule
        \multirow{5}{*}{FLUX} & \multirow{3}{*}{Prompt Opt.} & Baseline & 31.37 & 1.02 & 5.67 \\
        & & + Diffusion Probe (Ours) & 32.85 & \textbf{1.10} & 5.79 \\
        & & + LLM & \textbf{33.11} & \textbf{1.10} & \textbf{5.80} \\
        \cmidrule(lr){2-6}
        & \multirow{2}{*}{Seed Sel.} & Baseline (Random) & 31.37 & 1.02 & 5.67 \\
        & & + Diffusion Probe (Ours) & \textbf{31.41} & \textbf{1.06} & \textbf{5.79} \\
        \bottomrule
    \end{tabular}%
\end{table*}

\subsection{Experimental Details.}
\textbf{Models.}
Experiments are conducted on three prominent text-to-image models: Stable Diffusion XL (SDXL), FLUX.1-dev and Qwen-Image. SDXL and FLUX.1-dev for prompt optimization and seed selection and FLUX.1-dev for accelerating Flow-GRPO training. Prompt optimization and seed selection tasks use 25 inference steps. For Flow-GRPO training, the sampling steps are reduced from 6 to \textbf{2}, enabling significantly faster results. We test the performance of our Diffusion Probe on three obf the above models. All experiments are run on NVIDIA H100 80GB GPUs.

\textbf{Datasets}
Our experimental data is derived from the MS-COCO 2017 captions dataset. We constructed a training set of 15,000 unique prompts to train the Diffusion Probe. For evaluation, we curated a disjoint test set of 5,000 prompts. All subsequent experiments, including probe performance analysis and downstream tasks, are conducted exclusively on this unseen test set.

\textbf{Evaluation Metrics.}
We evaluate our framework on three axes: probe accuracy, downstream utility, and computational cost. To measure the probe’s predictive accuracy, we compute the correlation between its early-stage predictions and ground-truth scores using Spearman’s Rank Correlation (SRCC), Kendall’s Tau (KTC), Pearson Correlation (PCC), and the Area Under the ROC Curve (AUC-ROC). The practical utility in downstream tasks is then assessed by the quality of the final generated images, which we evaluate using CLIP Score~\cite{radford2021learningtransferablevisualmodels} for text-image alignment, ImageReward~\cite{xu2023imagereward} for human preference, and a general Aesthetic Score. Finally, we report computational efficiency in terms of trainable parameters, theoretical FLOPS, and wall-clock inference latency. We explain our metrics in detail in Appendix~\ref{sec:appendix_details_experiments}.

\textbf{Implementation Details.}
We configure the diffusion process for 25 total inference steps and design our Diffusion Probe to operate on cross-attention maps extracted at an early stage, specifically at step $t=5$. This extraction strategy is architecture-aware: for UNet-based models like SDXL, we collect maps from the final 10 encoder blocks, whereas for DiT-based models like FLUX, we use 10 consecutive blocks from the middle of the architecture. The choice of $t=5$ is justified by an ablation study detailed in Appendix~\ref{sec:appendix_diffusion_probe_results}. 

The probe is trained to predict the final image's ImageReward score, and to ensure robustness, our training procedure mitigates the natural data imbalance by strategically oversampling low-score instances. For downstream validation, we configure two tasks: Seed Selection, evaluating 10 distinct seeds per prompt, and Prompt Optimization, generating 4 prompt variations via the Qwen-3-Max API. In both scenarios, the probe's prediction facilitates an informed selection of the most promising candidate, thereby circumventing the need to run costly full generations for all options.

 While our downstream applications exclusively utilized probes trained with ImageReward, we observe a high correlation among various image quality assessment metrics. We further investigate the prediction accuracy of probes trained under diverse evaluation metrics as labels in Appendix~\ref{sec:appendix_diffusion_probe_results}.
 
\subsection{Diffusion Prober Performance Evaluation}
\begin{figure}
    \centering
    \includegraphics[width=\linewidth]{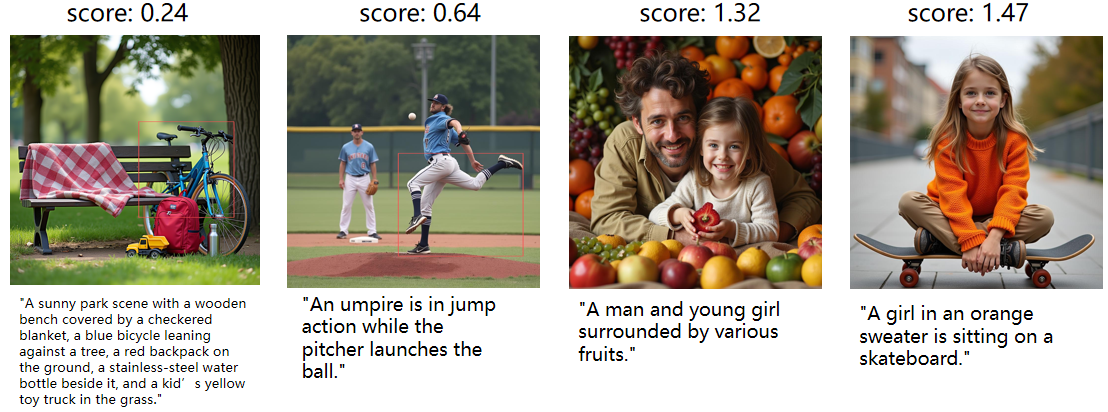}
    \caption{Diffusion Probe as an effective filter for aesthetically poor generations. By leveraging training on image quality labels, the probe accurately flags images with common defects (e.g., distorted anatomy, poor composition), assigning them low scores.}
    \label{fig:showcase}
\end{figure}

\textbf{Qualitative Results.}
Figure~\ref{fig:showcase} demonstrate that the Diffusion Probe’s numerical scores directly correspond to tangible visual quality and semantic correctness. We observed that images assigned low scores by our prober consistently exhibited clear failure modes, such as object omission and attribute mismatch. In stark contrast, high-scoring images were visually coherent and accurately reflected the prompt’s intent. This provides strong visual evidence that our Diffusion Probe has successfully learned to distinguish between successful generations and semantically flawed outcomes.

\textbf{Quantitative Results.}
As presented in Table~\ref{tab:comprehensive_results}, our Diffusion Probe demonstrates both high efficacy and broad generalizability across diverse model architectures. A clear trend emerges across all tested models: the probe’s predictive power is minimal at the initial step but increases substantially at an early-intermediate stage, consistently peaking around step 10 before plateauing. This indicates that a reliable quality signal materializes long before the generation process completes.

Quantitatively, the probe demonstrates remarkable predictive accuracy. On the state-of-the-art FLUX model, it yields peak scores by step 10, with an SRCC of \textbf{0.79}, AUC of \textbf{0.91}, KTC of \textbf{0.64}, and PCC of \textbf{0.78}. This strong performance is not confined to a single architecture; it is consistently replicated on the widely-used SDXL and the distinct Qwen-Image models, achieving high SRCC scores of \textbf{0.76} and \textbf{0.72}, respectively.

The consistently high correlation (SRCC, KTC, PCC) and classification (AUC-ROC) scores across these architecturally varied platforms robustly confirm that our method can reliably predict final image quality from early-stage signals. This validates our approach as a model-agnostic tool for optimizing generative applications. We also test the performance of Diffusion probe at 512$\times$512 resolution, please turn to Appendix~\ref{sec:appendix_diffusion_probe_results} for results. 

\subsection{Comprehensive Applications Results}

\textbf{Prompt Optimization:} In Table~\ref{tab:combined_results}, Diffusion Probe consistently yields substantial quality improvements over baseline methods for both SDXL and FLUX, achieving gains across CLIP Score, ImageReward, and Aesthetic Score. Notably, our lightweight approach attains performance competitive with much heavier LLM-based optimization methods, highlighting significant computational savings.

\textbf{Seed Selection:} Table~\ref{tab:combined_results} further demonstrates that Diffusion Probe effectively enhances output quality while dramatically reducing computational overhead. For the FLUX model, intelligently selected seeds elevate the Aesthetic Score from 5.67 to 5.79 and ImageReward from 1.02 to 1.06. This is achieved through a minimal pre-screening phase that circumvents computationally expensive, full-inference generations, thereby replacing an unguided, brute-force strategy with a resource-efficient, informed selection procedure.

\begin{figure}
    \centering
    \includegraphics[width=\linewidth]{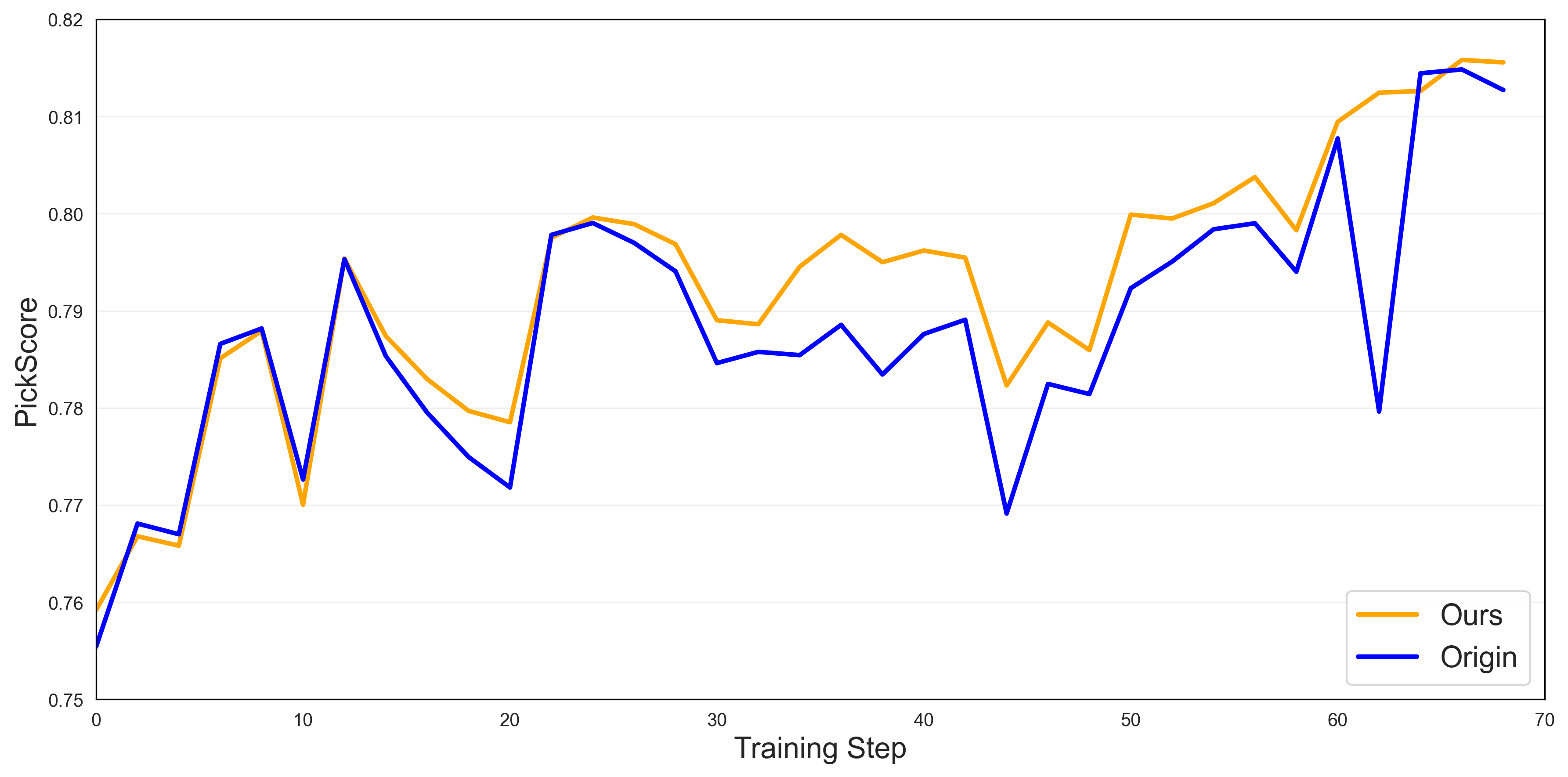}
    \caption{
        Comparison of the PickScore during training steps for our method ("Ours") versus the baseline method ("Origin") applied to Flow-GRPO. The plot demonstrates that our approach enhances the stability and convergence speed of the training process, as evidenced by the smoother fluctuations and faster rise in the PickScore across training steps. This indicates a more consistent and efficient learning process when using our method.
    }
    \label{fig:flowgrpo}
\end{figure}

\textbf{Efficient Flow-GRPO:} As shown in Figure~\ref{fig:flowgrpo}, integrating Diffusion Probe significantly enhances RL sample efficiency. By enriching training batches with 2.5$\times$ more high-quality samples , our method yields a markedly smoother and more stable convergence trajectory, as evidenced by the reward curves. This enhanced stability, stemming from cleaner gradient signals, accelerates policy convergence to the target reward. Consequently, this leads to substantial savings in both computational resources and development time. Please turn to Appendix~\ref{sec:appendix_flowgrpo} for detailed analysis.

\subsection{Ablation Study}

To validate the design choices of Diffusion Probe, we conduct a series of ablation experiments. We investigate the impact of two key hyperparameters: the diffusion step from which attention maps are collected and the spatial resolution of those maps.

\textbf{Impact of Target Step.}
The choice of early exiting step, $T_0$, for attention map extraction presents a critical trade-off: earlier steps maximize computational savings but risk noisy inputs, while later steps offer richer semantic information at reduced efficiency gains. To optimize this balance, we trained and evaluated Diffusion Probes for $T_0 \in \{1, 5, 10, 15\}$. As depicted in Table~\ref{tab:comprehensive_results}, prediction accuracy consistently improves with increasing $T_0$. Critically, we observe the most significant gains occur from $T_0=1$ to $T_0=5$, with marginal improvements thereafter. Consequently, $T_0=5$ was selected as our default, providing substantial predictive power at an extremely low computational overhead, representing an optimal accuracy-efficiency trade-off.

\subsection{Conclusion}
We introduced the \textbf{Diffusion Probe}, a lightweight model that predicts final image quality from early-stage cross-attention statistics. These early attention patterns provide a reliable predictive signal well before generation completes, and experiments across SDXL and FLUX confirm that the probe generalizes across model architectures. This capability improves the efficiency of \textbf{Prompt Optimization}, \textbf{Seed Selection}, and \textbf{Accelerated RL Training}: targeted sampling avoids full computation for low-potential candidates while improving final output quality. Diffusion Probe therefore offers a broadly applicable way to make modern T2I systems more efficient.

%% file: sec/X_suppl.tex
\clearpage
\setcounter{page}{1}
\maketitlesupplementary

\section{More Results about Diffusion Probe}
\label{sec:appendix_diffusion_probe_results}

Multimodal intelligence has also advanced rapidly across vision--language safety, voice-supervised scene-text understanding, visually grounded GUI agents, generative design, and virtual try-on~\cite{qiu2026yuvionvl,tang2022voice,zhai2026guide,niu2026cmecad,cui2026simpleposter,li2025shoefit}.

\begin{table*}[h!]
    \centering
    \caption{
        \textbf{Computational Cost Analysis.} This table compares the computational cost of naive brute-force workflows against our Diffusion Probe-guided approach. 'Single Generation' serves as the baseline, representing the cost of one complete image generation (14.70s), while the subsequent row highlights the negligible overhead of a single probe prediction (+0.05s). We evaluate two scenarios: a \textbf{10-candidate Seed Selection} and a \textbf{4-candidate Prompt Optimization}, demonstrating significant savings in both latency and FLOPS.
    }
    \label{tab:comprehensive_cost_analysis}
    \resizebox{\textwidth}{!}{ 
    \begin{tabular}{l l l r r}
        \toprule
        \textbf{Task} & \textbf{Workflow} & \textbf{Cost Breakdown} & \textbf{Total FLOPS (T)}  & \textbf{Total Latency (s)}  \\
        \midrule
        \multirow{2}{*}{Reference} & Single Generation & 1 $\times$ Full Gen. & 1877.56 & 14.70 \\
                                   & Single Gen. + 1 $\times$ Pred. & 1 $\times$ Full Gen. + 1 $\times$ Pred. & 1877.57 & 14.75 \\
        \midrule[0.8pt]
        \multirow{2}{*}{Seed Selection} & Naive Oversampling & 10 $\times$ Full Gen. & 18775.60& 147.00 \\
                                        & \textbf{Diffusion Probe Guided (Ours)} & \textbf{1 $\times$ Full Gen. + 10 $\times$ Pred.} & \textbf{5280.43} & \textbf{42.62} \\
        \midrule
        \multirow{2}{*}{Prompt Optimization} & Naive Comparison & 4 $\times$ Full Gen. & 7,510.25 & 58.00 \\
                                             & \textbf{Diffusion Probe Guided (Ours)} & \textbf{1 $\times$ Full Gen. + 4 $\times$ Pred.} & \textbf{3026.42} & \textbf{28.29} \\
        \bottomrule
    \end{tabular}
    }
\end{table*}

\subsection{The performance of Diffusion Probe at 512$\times$512 resoluton}

While our main experiments are conducted at a resolution of 1024$\times$1024, we additionally evaluate the Diffusion Probe at a lower resolution of 512$\times$512 to assess its robustness across input scales. Even under this reduced-resolution setting, the probe exhibits stable performance and remains well aligned with the target quality metrics, indicating that its predictive capability does not depend on the high-resolution regime. The corresponding results are presented in Table~\ref{tab:flux_near5_combined}.



\subsection{More Results of Ablation Study}

\textbf{Diffusion Probe trained with another metric.}
In the main text, the diffusion probe is trained using image–reward annotations as supervision. To further examine the probe’s flexibility, we additionally train it with several alternative image–quality metrics as labels. Across all cases, the probe is able to accurately approximate the corresponding quality indicators, demonstrating its robustness to different supervisory signals. The results are summarized in Table~\ref{tab:flux_step5_label_categories}.

\begin{table*}[h!]
    \centering
    \small
    \caption{
        Diffusion Probe's Predictive Accuracy for External Image Quality Metrics across Diffusion Steps near 5, evaluated on two resolutions (1024$\times$1024 and 512$\times$512). 
        This table reports the predictive alignment between the Diffusion Probe's internal scores and standard external image quality metrics under different input resolutions. Higher values indicate better predictive accuracy.
    }
    \label{tab:flux_near5_combined}
    \setlength{\tabcolsep}{4pt}
    \begin{tabular}{l l c | c c c c}
        \toprule
        \textbf{Base Model} & \textbf{Resolution} & \textbf{Step} & \textbf{SRCC $\uparrow$} & \textbf{AUC-ROC $\uparrow$} & \textbf{KTC $\uparrow$} & \textbf{PCC $\uparrow$} \\
        \midrule
        \multirow{8}{*}{\textbf{FLUX}}
            & \multirow{4}{*}{1024$\times$1024}
                & 3 & 0.69 & 0.83 & 0.53 & 0.68 \\
            &   & 4 & 0.75 & 0.88 & 0.60 & 0.74 \\
            &   & 6 & 0.77 & 0.90 & 0.62 & 0.70 \\
            &   & 7 & 0.74 & 0.87 & 0.58 & 0.72 \\
        \cmidrule(lr){2-7}
            & \multirow{4}{*}{512$\times$512}
                & 3 & 0.61 & 0.78 & 0.46 & 0.60 \\
            &   & 4 & 0.66 & 0.81 & 0.50 & 0.64 \\
            &   & 6 & 0.69 & 0.83 & 0.52 & 0.67 \\
            &   & 7 & 0.67 & 0.82 & 0.51 & 0.65 \\
        \bottomrule
    \end{tabular}
\end{table*}

\begin{table*}[h!]
    \centering
    \small
    \caption{
        Diffusion Probe's Predictive Accuracy for External Image Quality Metrics across Diffusion Step 5 (FLUX Model, 1024$\times$1024 Resolution) with Different Label Categories.
        This table reports the predictive alignment between the Diffusion Probe's internal scores and external image quality metrics, trained under two different label categories: Aesthetic Score and CLIP-SCORE. Higher values indicate better predictive accuracy.
    }
    \label{tab:flux_step5_label_categories}
    \setlength{\tabcolsep}{4pt}
    \begin{tabular}{l l c | l c c c c}
        \toprule
        \textbf{Base Model} & \textbf{Resolution} & \textbf{Step} & \textbf{Label Category} & \textbf{SRCC $\uparrow$} & \textbf{AUC-ROC $\uparrow$} & \textbf{KTC $\uparrow$} & \textbf{PCC $\uparrow$} \\
        \midrule
        \multirow{2}{*}{\textbf{FLUX}} 
            & \multirow{2}{*}{1024$\times$1024} 
            & 5  & \textbf{Aesthetic Score} & 0.74 & 0.82 & 0.58 & 0.73 \\
            &                                   & 5  & \textbf{CLIP-SCORE}      & 0.77 & 0.84 & 0.62 & 0.76 \\
        \bottomrule
    \end{tabular}
\end{table*}

\textbf{Ablation Study of Steps Window.}
We examine the effect of varying the diffusion steps window on the performance of the Diffusion Probe. Our primary goal is to demonstrate that the Diffusion Probe does not only exhibit strong predictive accuracy at the fifth diffusion step, but also performs well across a range of neighboring steps, indicating that the effective window extends beyond a single step. Specifically, we analyze the probe's ability to predict image quality metrics at steps in the vicinity of step 5, such as steps 3, 4, 6, and 7.

By systematically evaluating the model across these steps, we aim to show that the probe captures relevant image quality features consistently over a broader set of diffusion stages, rather than relying solely on step 5. This extended effective window suggests that the probe's attention features are stable and robust, maintaining high predictive accuracy across multiple diffusion steps.

As shown in Table~\ref{tab:flux_near5_combined}, the Diffusion Probe maintains similar levels of predictive accuracy across the steps near step 5, with only marginal variations in performance. This result confirms that the probe remains effective over a range of neighboring steps, highlighting the flexibility and robustness of our method in capturing quality-related features at different stages of the diffusion process.

\subsection{Results of Efficient Flow-GRPO}
\label{sec:appendix_flowgrpo}
In this section, we present the empirical results of integrating our \textbf{Diffusion Probe} into the Flow-GRPO framework, demonstrating the practical benefits of our early-stage quality predictor in a resource-intensive application. Our evaluation focuses on two key aspects of efficiency that are directly impacted by the probe's filtering capability:

\begin{itemize}
    \item \textbf{Training Efficiency Improvement:} We quantify the acceleration of the overall training process, measuring the reduction in computational time and resources required to achieve comparable or superior model performance.
    
    \item \textbf{Increase in Effective Sample Ratio:}  
    We analyze how our Diffusion Probe enhances the sample efficiency during the training loop, specifically within the Flow-GRPO framework. The sampling process in Flow-GRPO aims to capture both positive and negative samples, encouraging diversity and a balanced exploration of the reward space. In this context, we measure the variance of the Pick Score across the sampled data at each training step. A higher variance in the Pick Score indicates a better distinction between positive and negative samples, which is crucial for training stability and performance. By using our probe to filter out low-quality samples early in the process, the variance of the Pick Score becomes more focused on high-quality, informative samples. As a result, the proportion of valid training data increases by 40\%, reflecting a significant improvement in the sample efficiency and diversity, without sacrificing the effectiveness of positive-negative sample separation.

\end{itemize}

Figure~\ref{fig:flowgrpo} that our approach not only accelerates the training pipeline but also makes it more effective by optimizing the data used for policy updates.

\subsection{More Qualitative Results about Diffusion Probe}

\begin{figure*}
    \centering
    \includegraphics[width=\linewidth]{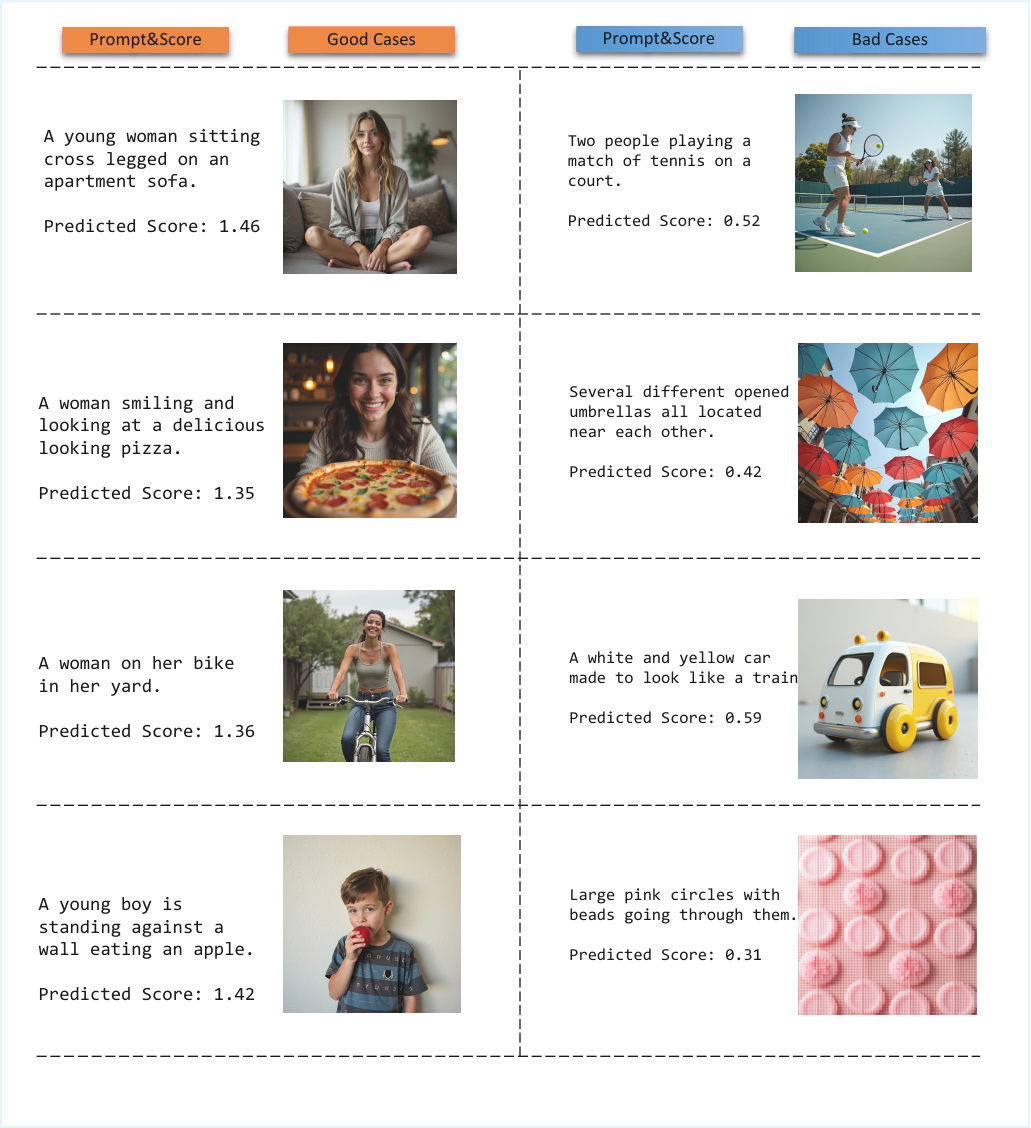}
    \caption{
    More qualitative results about Diffusion Probe
    }
    \label{fig:attribute_normalcase}
\end{figure*}

In Figure~\ref{fig:attribute_normalcase}, we provide additional detailed qualitative examples of our Diffusion Probe, showcasing its ability to evaluate images generated at different quality levels. We rank 10 images generated from different prompts and compare these rankings with the corresponding original image quality metrics. This comparison highlights the Diffusion Probe’s effectiveness in assessing both the visual consistency between the text prompt and the generated image, as well as the overall aesthetic quality of the images.

Furthermore, we demonstrate how well the Diffusion Probe’s rankings align with existing image quality metrics, such as those based on human perception. The results illustrate that the Diffusion Probe can effectively capture both text-image consistency and image appeal, making it a reliable tool for quality assessment. The alignment with established metrics further validates the probe’s predictive accuracy, emphasizing its potential as a robust quality evaluator for generated images.

\subsection{Computation Cost Analysis}
\label{sec:appendix_computationcost}
We show the computation cost in Table~\ref{tab:comprehensive_cost_analysis}. A hallmark of Diffusion Probe is its exceptional computational efficiency. As detailed in Table~\ref{tab:comprehensive_cost_analysis}, a single probe prediction requires only 0.0036 TFLOPS and 0.05s—orders of magnitude less than the 1877.56 TFLOPS and 14.70s demanded by a full generation.

This efficiency enables dramatic accelerations in real-world workflows. For a 10-candidate Seed Selection task, our probe-guided approach slashes latency from 147.00s to 42.62s, resulting in a \textbf{3.45$\times$} speedup. Similarly, in a 4-candidate Prompt Optimization task, it reduces latency from 58.00s to 28.29s, yielding a \textbf{2.05$\times$} speedup.

By strategically substituting expensive full-generation rollouts with near-instantaneous probe predictions for all but the final candidate, Diffusion Probe provides substantial computational and temporal savings, validating its role as a practical and powerful tool for optimizing large-scale generative workflows.

\begin{table}[h]
    \centering
    \footnotesize 
    \renewcommand{\arraystretch}{1.0}
    \setlength{\tabcolsep}{3.2pt} 
    \caption{\textbf{Robustness across Architectures, Resolutions, and Steps.} Our probe is fixed (trained on FLUX). Results across IR and CLIP metrics.}
    \label{tab:arch_step}
    \begin{tabular}{l c c c l | cccc}
        \toprule
        \textbf{Model} & \textbf{Res.} & \textbf{$N$} & \textbf{Ext. $t$} & \textbf{Met.} & \textbf{SRCC} & \textbf{AUC} & \textbf{PCC} & \textbf{KTC} \\ 
        \midrule
        SDXL & 768 & 25 & 5 & IR & 0.70 & 0.81 & 0.51 & 0.69 \\
        SDXL & 768 & 25 & 5 & \textbf{CLP} & \textbf{0.63} & \textbf{0.74} & \textbf{0.42} & \textbf{0.61} \\
        \midrule
        \multirow{5}{*}{\textbf{FLUX}} & \multirow{5}{*}{2048} & \multirow{5}{*}{25} & \textbf{5} & IR & \textbf{0.79} & \textbf{0.91} & \textbf{0.64} & \textbf{0.76} \\
                                     & & & \textbf{5} & \textbf{CLP} & \textbf{0.72} & \textbf{0.82} & \textbf{0.56} & \textbf{0.71} \\
                                     & & & 21 & IR & 0.76 & 0.88 & 0.61 & 0.72 \\
                                     & & & 22 & IR & 0.75 & 0.89 & 0.60 & 0.73 \\
                                     & & & 24 & IR & 0.74 & 0.88 & 0.59 & 0.70 \\
        \bottomrule
    \end{tabular}
\end{table}

\begin{table}[h]
    \centering
    \footnotesize
    \renewcommand{\arraystretch}{1.0}
    \setlength{\tabcolsep}{3.5pt} 
    \caption{\textbf{Generalization and Human Alignment.} Zero-shot performance across prompt complexities and sampling steps.}
    \label{tab:qwen_study}
    \begin{tabular}{l c c c | cccc | c}
        \toprule
        \textbf{Prompt} & \textbf{Res.} & \textbf{$N$} & \textbf{Ext. $t$} & \textbf{SRCC} & \textbf{AUC} & \textbf{PCC} & \textbf{KTC} & \textbf{User} \\
        \midrule
        \multirow{3}{*}{Simple}  & \multirow{3}{*}{2048} & 10 & 2  & 0.73 & 0.82 & 0.71 & 0.58 & 79.2\% \\
                                 &                       & 25 & 5  & 0.72 & 0.87 & 0.71 & 0.56 & 83.5\% \\
                                 &                       & 50 & 10 & 0.74 & 0.88 & 0.72 & 0.57 & 81.3\% \\
        \midrule
        \multirow{3}{*}{Complex} & \multirow{3}{*}{2048} & 10 & 2  & 0.69 & 0.79 & 0.67 & 0.53 & 76.0\% \\
                                 &                       & 25 & 5  & 0.68 & 0.81 & 0.66 & 0.52 & 79.1\% \\
                                 &                       & 50 & 10 & 0.70 & 0.83 & 0.68 & 0.54 & 77.5\% \\
        \bottomrule
    \end{tabular}
\end{table}

\subsection{Robustness and Generalization Analysis}
\label{subsec:robustness}

To verify the reliability of our proposed probe, we conduct a comprehensive robustness analysis across various dimensions, including model architectures, sampling trajectories, and semantic complexity. Although trained exclusively on \textbf{FLUX}, the following results demonstrate its exceptional zero-shot generalization.

\textbf{Universal Generalization across Architectures and Steps.} 
As detailed in Tab.~\ref{tab:arch_step} and Tab.~\ref{tab:qwen_study}, the probe exhibits significant structural and temporal invariance. It maintains \textbf{consistent robustness} across diverse backbones and pixel densities, capturing universal quality signals rather than over-fitting to model-specific artifacts. Notably, the performance remains stable across different sampling budgets (\textbf{10, 25, or 50 steps}) at equivalent noise levels. This \textit{step-invariance} proves that the probe relies on SNR-linked features rather than specific sampling trajectories. Furthermore, without fine-tuning, the probe generalizes to \textbf{CLIP Score (SRCC 0.72)}, effectively distilling universal features such as aesthetic quality and semantic alignment.

\textbf{Architectural Insights and Prompt Complexity.} 
Our design choices are grounded in structural necessity. Ablation studies show that cross-attention significantly outperforms self-attention (\textbf{SRCC 0.76 vs. 0.61}) in capturing semantic-structural alignment. The probe also sustains high performance (\textbf{AUC $> 0.78$}) even when processing long, multi-object prompts. When filtering by specific Parts-of-Speech (POS) tags (e.g., nouns and adjectives), accuracy drops by \textbf{only $\sim$10\%}, suggesting a reliance on holistic context rather than isolated keywords. Additionally, a \textbf{three-head design} maintains high accuracy (\textbf{SRCC $\approx$ 0.76}), enabling multi-dimensional assessment of both alignment and aesthetics.

\textbf{Efficiency vs. Trajectory Robustness.} 
The probe provides valid predictive power as early as \textbf{Step 3/25}, while remaining accurate (\textbf{SRCC $> 0.70$}) for artifacts emerging late in the trajectory (e.g., $t \in [21, 24]$). Identifying these ``bad cases'' early \textbf{drastically reduces inference costs} by avoiding redundant decoding. Regarding image diversity, the number of semantic clusters ($N_{cls}$) remains stable post-optimization (\textbf{5.3 $\rightarrow$ 5.2}), confirming that our method prunes low-quality samples while \textbf{preserving generative variety} without inducing mode collapse.

\textbf{Correlation with Human Perception.} 
Finally, our \textbf{user study} (Tab.~\ref{tab:qwen_study}) confirms a \textbf{74\% agreement} with human preferences. This validates the probe as a \textbf{reliable perceptual proxy} for real-world generative applications, ensuring that the automated quality assessment aligns with actual human judgment.

\subsection{More failure modes of generated images and corresponding cross attention maps}
\label{sec:appendix_failure}

\begin{figure*}
    \centering
    \includegraphics[width=0.7\linewidth]{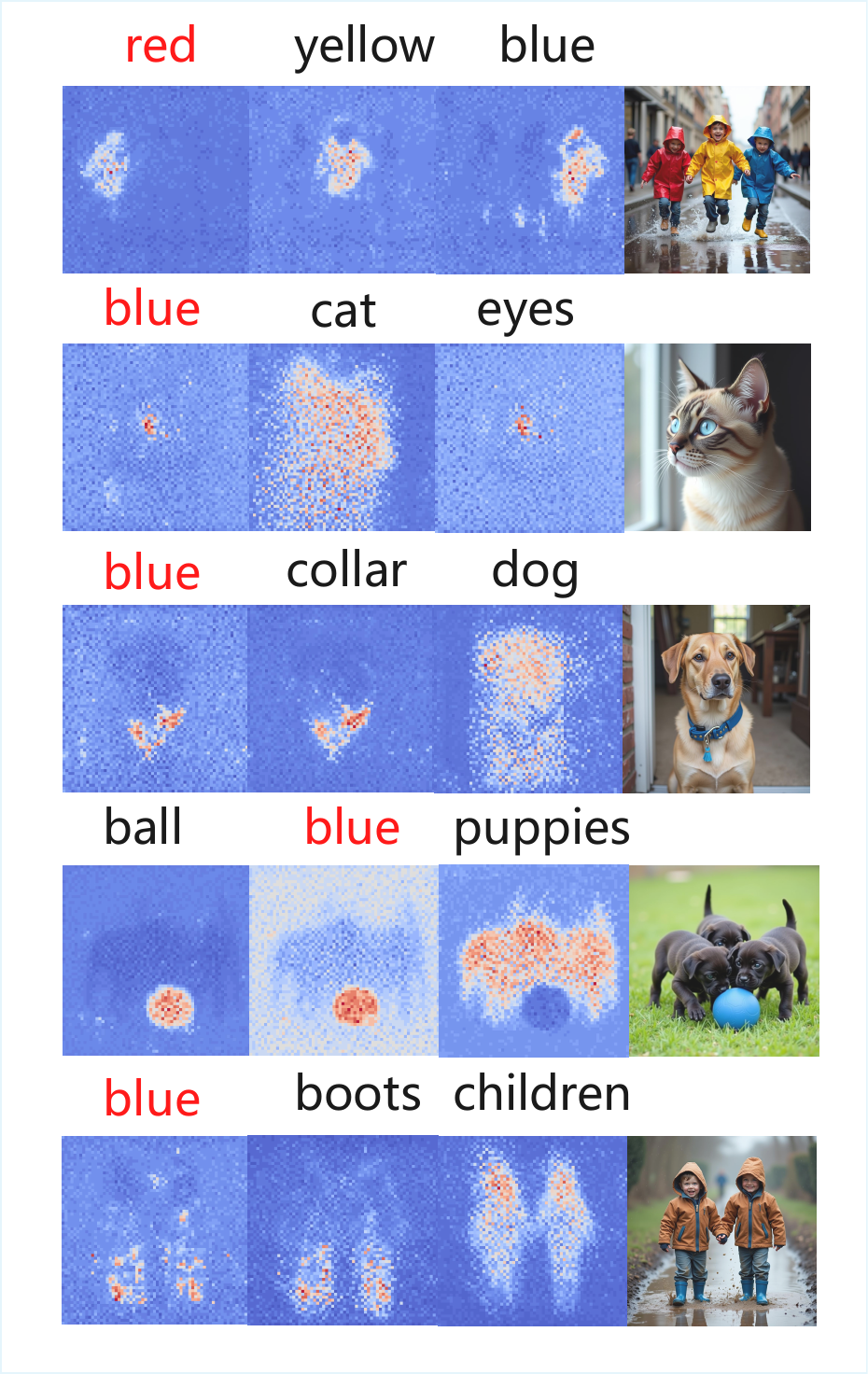}
    \caption{
        Normal case illustrating generated images and their corresponding cross-attention maps when the image generation is successful with no quality issues. In these cases, the cross-attention maps are well-focused, highlighting the specific regions of the image that correspond to the key features of the prompt. This indicates that when the generated image quality is high, the attention mechanism remains concentrated on the relevant visual attributes, reflecting the model's proper alignment with the textual description.
    }
    \label{fig:attribute_normalcase2}
\end{figure*}

\begin{figure*}
    \centering
    \includegraphics[width=0.7\linewidth]{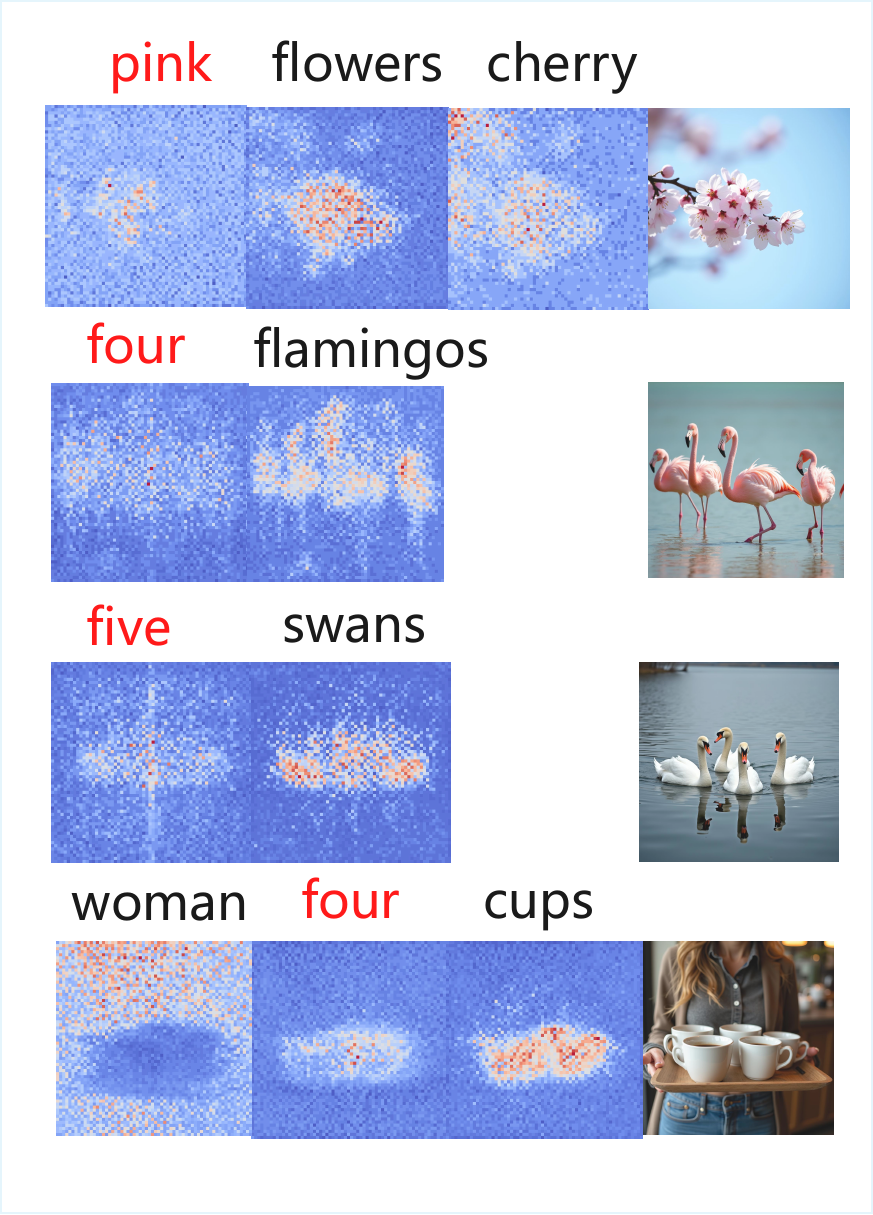}
    \caption{
        We list some cases of generation failures. When the generated image has attributes that do not match the prompt, the corresponding cross-attention map visualization exhibits a dissipation characteristic.
    }
    \label{fig:attribute_normalcase1}
\end{figure*}

As shown in Figure~\ref{fig:attribute_normalcase1} and Figure~\ref{fig:attribute_normalcase2}. In addition to the examples discussed in the main text, we further illustrate two representative failure modes of generated images: attribute mismatch and quantity mismatch. For attribute-mismatch cases, we find that the cross-attention maps corresponding to the specific attribute tokens become noticeably diffuse, suggesting that the model is unable to localize the intended attribute to the correct visual regions. 

In contrast, quantity-mismatch cases exhibit different attention behaviors. When the model generates fewer objects than specified, the attention associated with the quantity or category tokens typically collapses onto a single region, indicating a failure to distribute attention across multiple instances. Conversely, when the model produces more objects than required, the attention maps often become fragmented into several weak hotspots. These patterns highlight how deviations in attention allocation correlate with different forms of generation errors.

\section{Details about the experiments}
\label{sec:appendix_details_experiments}
\subsection{Details about our metric}

To quantitatively assess the performance of our \textbf{Diffusion Probe}, we follow a carefully controlled evaluation pipeline to compute the correlation and classification metrics. The procedure ensures both fairness and reproducibility.

\begin{enumerate}
    \item \textbf{Ground-Truth Data Preparation:}  
    For a given set of prompts, we first generate a large collection of final images using the base diffusion model. Each rendered image is then evaluated using a pre-trained aesthetic scoring model (e.g., the LAION aesthetic predictor) to obtain the ground-truth quality score \(S_{gt}\).  
    \textbf{To avoid distributional collapse in the test set—where scores might cluster heavily within a narrow interval—we deliberately adjust the selection of test samples so that the resulting distribution of \(S_{gt}\) spans a broad range of quality levels, including both low- and high-quality images.}  
    This ensures that evaluation metrics are not biased toward any particular score region.

    \item \textbf{Probe Prediction:}  
    During the generation process of the same set of images, our Diffusion Probe is activated at an early stage (typically within the first 10–20\% of denoising steps). The probe analyzes intermediate attention features and outputs a predictive score \(S_{probe}\) for each instance.

    \item \textbf{Metric Computation:}  
    Given the paired scores \((S_{probe}, S_{gt})\), we compute the evaluation metrics as follows:
    \begin{itemize}
        \item \textbf{SRCC and KTC:}  
        We compute the Spearman Rank Correlation Coefficient (SRCC) and Kendall Tau Coefficient (KTC) between the vectors of all \(S_{probe}\) and \(S_{gt}\) values. These metrics measure the consistency of the probe’s predicted ranking with the ground-truth ranking.

        \item \textbf{AUC-ROC:}  
        To assess classification performance, we binarize the ground-truth scores using the median of all \(S_{gt}\) values as the threshold. Images above the median are labeled as high-quality (class 1), and those below as low-quality (class 0). Using these binary labels and the probe's continuous predictions \(S_{probe}\), we compute the Area Under the ROC Curve (AUC-ROC), which reflects the probe’s ability to separate high- and low-quality samples.
    \end{itemize}
\end{enumerate}

This controlled process—especially the balanced construction of the ground-truth distribution—ensures that our evaluation reliably reflects the predictive capability of the Diffusion Probe across a wide spectrum of image qualities.
\clearpage

\begin{figure*}
    \centering
    \includegraphics[width=\linewidth]{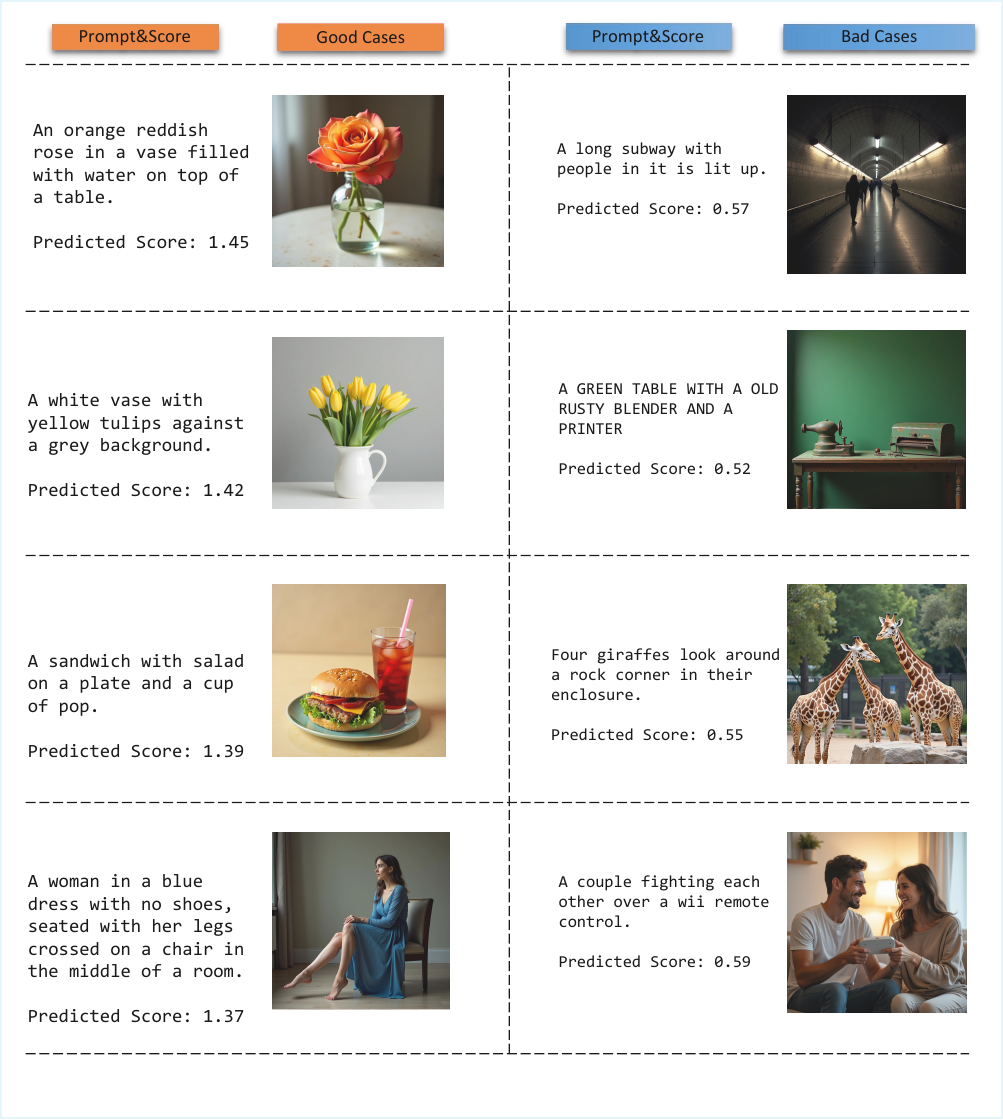}
    \caption{
    More qualitative results about Diffusion Probe
    }
    \label{fig:attribute_normalcase3}
\end{figure*}